%% file: main.tex
\documentclass{article}

\usepackage[preprint]{neurips_2025}

\usepackage[utf8]{inputenc} 
\usepackage[T1]{fontenc}    
\usepackage{hyperref}       
\usepackage{url}            
\usepackage{booktabs}       
\usepackage{amsfonts}       
\usepackage{nicefrac}       
\usepackage{microtype}      
\usepackage{bbm}
\usepackage{caption}
\usepackage{subcaption}
\usepackage{wrapfig}
\usepackage{multirow}
\usepackage{comment}
\usepackage{enumitem}
\usepackage{pifont}
\usepackage[pdftex]{graphicx}

\usepackage{nicefrac}         
\usepackage{microtype}        
\usepackage{bbm}              
\usepackage[utf8]{inputenc}   
\usepackage[T1]{fontenc}      

\usepackage{amsmath}
\usepackage{enumitem,amssymb}
\newlist{todolist}{itemize}{2}
\setlist[todolist]{label=$\square$}
\PassOptionsToPackage{numbers, compress}{natbib}

\newcommand{\ourcolor}[1]{\cellcolor{gray!25} #1}

\usepackage{amsthm}

\newtheorem{lemma}{Lemma}

\usepackage{amsmath, amssymb, amsthm}
\usepackage[ruled,vlined]{algorithm2e}
\SetKwInput{KwInput}{Input}
\SetKwInput{KwOutput}{Output}

\usepackage[dvipsnames]{xcolor}
\usepackage{colortbl}

\title{MultLFG: Training-free Multi-LoRA composition using Frequency-domain Guidance}

\author{Aniket Roy$^{1}$, Maitreya Suin$^{2}$, Ketul Shah$^{1}$, Rama Chellappa$^{1}$  \\
     Johns Hopkins University$^{1}$, Samsung AI Center Toronto$^{2}$
      }

\begin{document}
\maketitle
\input{Sec/0_abstract.tex}
\input{Sec/1_intro}
\input{Sec/2_related_work}
\input{Sec/2.5_preliminaries}
\input{Sec/motivation}

\input{Sec/3_method}
\input{Sec/theory_5}

\input{Sec/4_experiments}

\input{Sec/5_conclusion}
\bibliographystyle{plainnat}
\bibliography{name}


\end{document}

%% file: Sec/0_abstract.tex
\begin{abstract}
Low-Rank Adaptation (LoRA) has gained prominence as a computationally efficient method for fine-tuning generative models, enabling distinct visual concept synthesis with minimal overhead. However, current methods struggle to effectively merge multiple LoRA adapters without training, particularly in complex compositions involving diverse visual elements. We introduce MultLFG, a novel framework for training-free multi-LoRA composition that utilizes frequency-domain guidance to achieve adaptive fusion of multiple LoRAs. Unlike existing methods that uniformly aggregate concept-specific LoRAs, MultLFG employs a timestep and frequency subband adaptive fusion strategy, selectively activating relevant LoRAs based on content relevance at specific timesteps and frequency bands. This frequency-sensitive guidance not only improves spatial coherence but also provides finer control over multi-LoRA composition, leading to more accurate and consistent results. Experimental evaluations on the ComposLoRA benchmark reveal that MultLFG substantially enhances compositional fidelity and image quality across various styles and concept sets, outperforming state-of-the-art baselines in multi-concept generation tasks. Code will be released.
\end{abstract}

%% file: Sec/1_intro.tex
\vspace{-0.5cm}
\section{Introduction}
\vspace{-0.3cm}

In recent developments within text-to-image synthesis \cite{ramesh2022hierarchical,saharia2022image, rombach2022high, esser2021taming}, Low-Rank Adaptation (LoRA) \cite{hu2022lora} has emerged as a highly efficient method for fine-tuning generative models with minimal computational cost. By leveraging LoRA, we can achieve fine-grained control over the generated content, enabling the production of precise visual elements such as specific characters, tailored garments, unique artistic styles, and other notable features that enhance the fidelity of the output images. 
However, while individual LoRA modules demonstrate significant promise, existing techniques encounter difficulties when attempting to combine multiple LoRAs into a single generative process, particularly as their number increases, which complicates the creation of intricate, composite images. This limitation naturally leads to the question: How can we seamlessly merge several pre-trained LoRAs in a training-free manner while preserving the distinct attributes that each module contributes to the overall image?

\begin{figure}
    \centering
    \includegraphics[width=12cm]{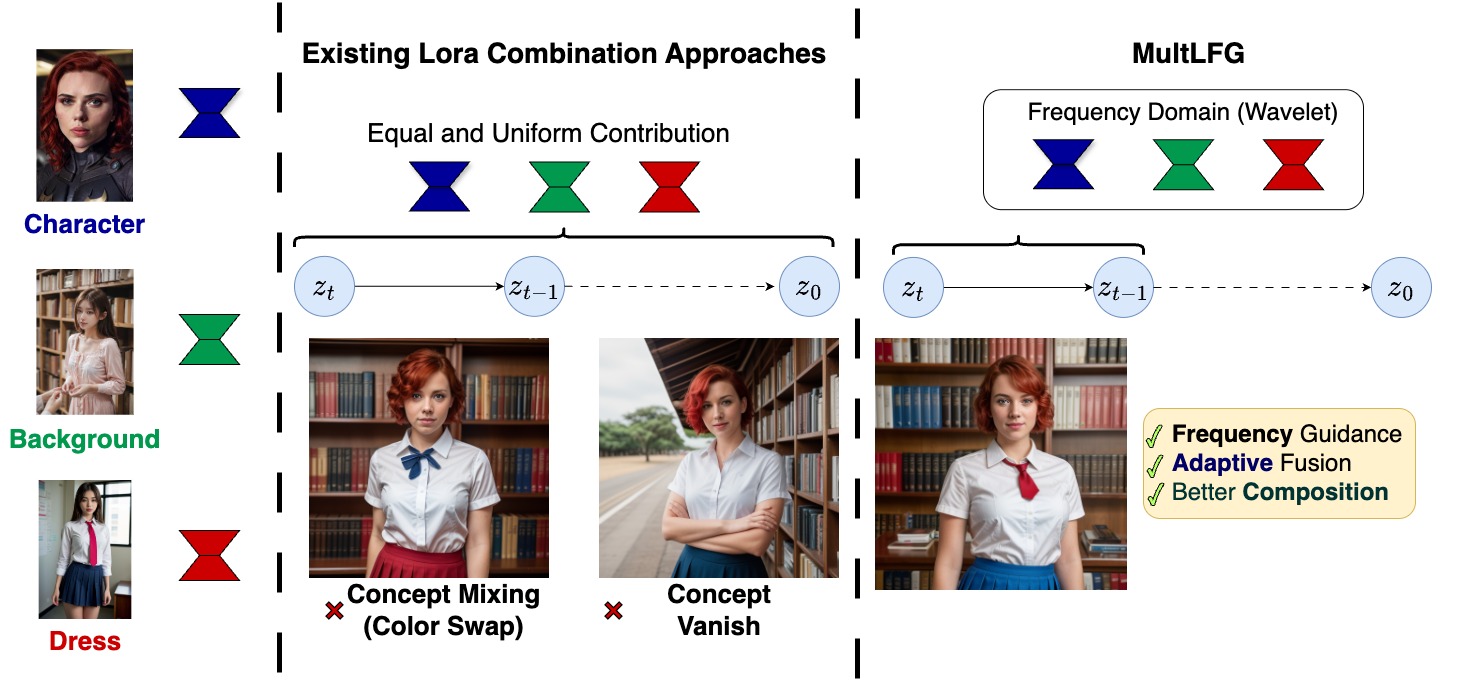}
    \vspace{-0.2cm}
    \caption{\small{Existing LoRA composition methods (composite, switch~\cite{zhong2024multi}) generally uses equal and uniform contribution of each concept LoRAs across denoising timesteps, incurring concept mixing or erasure. MultLFG performs multi-LoRA composition using Wavelet based frequency guidance and adaptive fusion providing better compositionality.}}
    \label{fig:teaser}
    \vspace{-0.5cm}
\end{figure}

Merging multiple concepts without additional training is a complex challenge, requiring both precise spatial localization of each concept and seamless blending of their features. Existing methods like LoRA-composite~\cite{zhong2024multi} and LoRA-switch~\cite{zhong2024multi} attempt to aggregate concepts by averaging or switching LoRAs across timesteps but fail to enforce spatial consistency. This often results in misplaced objects, concept vanish or concept mixing, as seen in Fig.~\ref{fig:teaser}, where the tie is sometimes missing or it's color (red) is interchanged with the skirt color (blue). These issues arise due to attributing equal weight to all LoRAs irrespective of their importance at a particular timestep or solely operating in the spatial domain, which is inadequate for handling fine-grained or complex compositions. To address this, we propose a frequency-based guidance mechanism with adaptive weights, enabling more controlled merging of concepts across frequency sub-bands. While previous work~\cite{zou2025cached} employed Fourier-based caching to group LoRAs into low and high-frequency bands, it overlooked the multi-scale nature of concept details, such as backgrounds (bookshelf) containing both high-frequency text and low-frequency surfaces (Fig.~\ref{fig:teaser}). The caching approach also involves frequency-based grouping that relies on hyperparameters, making it less scalable.

To address these limitations, we leverage wavelets~\cite{mallat1989theory}, which decompose images into multi-scale subbands, enabling finer control over concept composition. We observe that low-frequency components tend to dominate in the initial timesteps, while high-frequency details become more pronounced during later stages of denoising. A greater change relative to previous timesteps indicates a stronger influence of the corresponding LoRA module on the overall composition for that particular timestep. These observations motivate the design of MultLFG, a multi-LoRA composition framework that is both frequency-subband and timestep adaptive. MultLFG automatically identifies relevant content across timesteps and frequency bands for generation, selectively activating the appropriate LoRAs. This frequency-sensitive guidance framework not only enhances spatial coherence but also offers more precise control over multi-LoRA composition, resulting in improved consistency and compositional fidelity.



Our major contributions are as follows:
\begin{itemize}
    \item We introduce MultLFG - a framework for training-free multi-lora composition using frequency guidance and adaptive weighting.
    \item Instead of spatial composition, we propose a novel frequency domain guidance, which decomposes the composition into a wavelet-based frequency domain and performs frequency guidance in each of the frequency subbands. 
    \item We also propose an adaptive score weighting mechanism based on the contributions of each concept LoRAs for individual timesteps. This basically provides a importance-based weighted average of the concepts LoRAs during inference.
    \item We experiment on the ComposLoRA benchmark and obtain performance gains across different styles and number of LoRAs w.r.t CLIPScore, GPT-4v evaluation and user study.
\end{itemize}

%% file: Sec/2_related_work.tex
\vspace{-0.3cm}
\section{Related Work}
\vspace{-0.3cm}

\textbf{Multi-concept text-to-image generation.}
The ability to generate images that seamlessly blend multiple concepts is fundamental for tailoring digital content to meet specific user requirements. Prior work in this area has predominantly explored two strategies. One approach refines the diffusion process of generative models to more accurately adhere to given specifications \cite{jiang2024mc, kumari2022multi, xiao2024fastcomposer}, while the other assembles a collection of independent modules that enforce predefined constraints \cite{kwon2024concept, gu2023mix, zhong2024multi}.
Despite the success of conventional methods at producing images with broad thematic elements, they often fall short when it comes to the precise integration of user-defined objects \cite{kumari2022multi}. Meanwhile, techniques that do focus on incorporating specific objects typically demand extensive fine-tuning and struggle to handle multiple objects concurrently~\cite{shah2024ziplora, huang2023lorahub, roy2025duolora, borse2024foura, agarwal2024training, frenkel2024implicit, ouyang2025k, zhao2024merging, prabhakar2024lora, wu2024mixture, marta2025flora, chen2024iteris, liuhico}.

\textbf{Multi-LoRA compositions}
Recent efforts have increasingly explored the use of large language models (LLMs) and diffusion models as foundational architectures to fine-tune LoRA weights for a variety of purposes. These efforts aim to achieve tasks such as combining different visual elements in image synthesis \cite{shah2024ziplora}, reducing the parameter count for multi-modal inference \cite{chen2024llava, chavan2023one}, and tailoring models to specific application domains \cite{zhang2023composing, kong2024lora, li2024instruction}. In terms of LoRA composition strategies, LoraHub \cite{huang2023lorahub} employ few-shot demonstrations to learn coefficient matrices that effectively merge several LoRA modules into a single, unified module. Meanwhile, LoRA Merge \cite{zhong2024multi} relies on arithmetic operations, such as addition and negation, to combine LoRA weights. In contrast to these weight-based fusion techniques, approaches such as LoRA Switch and LoRA Composite \cite{zhong2024multi} preserve the original LoRA weights and instead modulate the interactions among them during the inference process. \cite{zou2025cached} propose to use frequency based caching for multi-lora composition.

%% file: Sec/2.5_preliminaries.tex
\vspace{-0.3cm}
\section{Preliminaries}
\vspace{-0.3cm}

\textbf{Diffusion models.} Diffusion models iteratively corrupt an image \( x_0 \) through Gaussian noise addition defined as \( q(x_t \mid x_{t-1}) = \mathcal{N}\bigl(x_t; \sqrt{1-\beta_t}\,x_{t-1}, \beta_t I\bigr) \)~\cite{dhariwal2021diffusion}. The state at any timestep \( t \) can be expressed as:
\[
x_t = \sqrt{\bar{\alpha}_t}\,x_0 + \sqrt{1-\bar{\alpha}_t}\,\epsilon_t, \quad \epsilon_t \sim \mathcal{N}(0, I)
\]
where \( \bar{\alpha}_t = \prod_{s=1}^t (1 - \beta_s) \). The reverse process approximates the posterior using a network \( \epsilon_\theta \) as:
\begin{align*}
   p_\theta(x_{t-1} \mid x_t) = \mathcal{N}\Bigl(x_{t-1}; \mu_\theta(x_t, t), \sigma_t^2 I\Bigr), \quad  
   \mu_\theta(x_t, t) = \frac{1}{\sqrt{\alpha_t}} \left( x_t - \frac{\beta_t}{1-\alpha_t}\,\epsilon_\theta(x_t, t) \right).
\end{align*}

During inference, to incorporate conditional information \( c \), classifier-free guidance adjusts the score:
\[
\hat{\epsilon}_\theta(x_t, c, t) = \epsilon_\theta(x_t, t) + \gamma\Bigl(\epsilon_\theta(x_t, c, t) - \epsilon_\theta(x_t, t)\Bigr)
\]
allowing for adaptive control of semantic contributions during generation.

\textbf{Discrete Wavelet Transform.}
The wavelet transform is a traditional method extensively applied in image compression to separate an image into its smooth, low-frequency components and its detailed, high-frequency components~\cite{mallat1989theory}. In this process, the low-frequency subbands (LL) resemble a reduced-resolution version of the original image, while the high-frequency subbands (HL, LH, HH) capture the fine local features such as vertical, horizontal, and diagonal edges (Fig.~\ref{fig:main_arch}). Among various wavelet techniques, the Haar wavelet is particularly favored in practical applications for its simple implementation. The procedure involves two key operations: the Discrete Wavelet Transform (DWT) and the Inverse Discrete Wavelet Transform (IDWT).

%% file: Sec/motivation.tex
\vspace{-0.3cm}
\section{Motivation}
\vspace{-0.3cm}
\begin{wrapfigure}{r}{0.45\textwidth}
    \centering
    \vspace{-45pt}
    \includegraphics[width=0.45\textwidth]{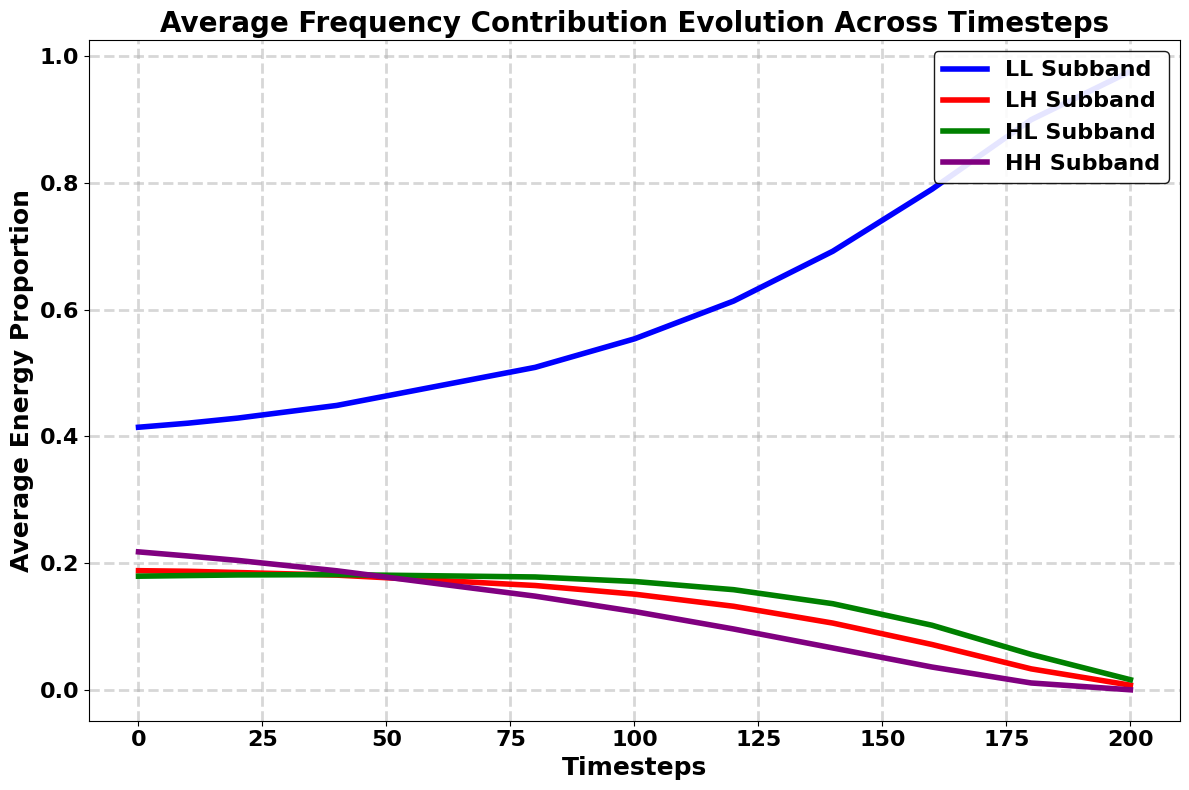}
    \vspace{-15pt}
    \caption{\small{Frequency analysis. Low-frequency components (LL) are prominant in early timesteps (T=200), whereas high-frequency components (HL, LH, HH) are more prominant in later timesteps (T=0) during denoising.}} 
    \label{fig:freq_analysis}
    \vspace{-20pt}
\end{wrapfigure}
Compared to LoRA-composite~\cite{zhong2024multi}, where at each timesteps, all the concepts are treated equally, we observe that the importance of concepts varies across timesteps and frequency subbands.

\textbf{Observation:} In the early stages of the denoising process, low-frequency components—corresponding to coarse structures and global shapes—tend to dominate. As denoising progresses toward later timesteps, high-frequency components—capturing fine-grained details and textures—become increasingly prominent.

Understanding the temporal evolution of frequency components is crucial for designing more effective guidance and composition strategies, especially in multi-concept generation tasks. If low-frequency information (e.g., background, layout, object silhouette) is established early and high-frequency information (e.g., texture, fine-grained attributes) appears later, this suggests that guidance signals or LoRA adapters could be selectively weighted across timesteps and frequency bands for better compositional control.

To verify this hypothesis, we perform a frequency-domain energy analysis across denoising timesteps. For a given prompt, we generate intermediate latent states $\{\mathbf{z}_t\}_{t=0}^{T}$ at each denoising timestep $t$ and decode them into images $\{\mathbf{x}_t\}$ using a pretrained VAE decoder.
Then, for each image $\mathbf{x}_t$, we apply DWT to obtain four subbands: $\text{DWT}(\mathbf{x}_t) \rightarrow \{x_t^{LL}, x_t^{LH}, x_t^{HL}, x_t^{HH}\}$,
where $x_t^{LL}$ captures the low-frequency structure and $\{x_t^{LH}, x_t^{HL}, x_t^{HH}\}$ represent high-frequency details (Fig.~\ref{fig:main_arch}).
Now, for each subband $b \in \{LL, LH, HL, HH\}$ at each timestep $t$, we compute the normalized energy, $E_t^b = \frac{\|x_t^b\|_2^2}{\sum_{b'} \|x_t^{b'}\|_2^2}$, which reflects the relative contribution of each frequency band. Finally, we plot $E_t^b$ as a function of timestep $t$ to observe how frequency contributions evolve during the reverse diffusion process in Fig.~\ref{fig:freq_analysis}.
Here, we generate around 3000 images for prompts in ComposeLoRA benchmark, and report the average wavelet analysis in Fig.~\ref{fig:freq_analysis}. 
We observe that the low-frequency subband ($LL$) dominates in early timesteps (high $E_t^{LL}$ for large $t$). The high-frequency subbands ($LH$, $HL$, $HH$) become more prominent in later timesteps (increasing $E_t^b$ as $t \to 0$), thereby validating the coarse-to-fine generation pattern of diffusion models. 
To maintain consistency with the terminology, timesteps are defined such that $t=200$ represents the initial timesteps, while $t=0$ indicates the later timesteps during denoising. These observation align with frequency analysis performed in Fourier-frequency domain by prior works~\cite{zou2025cached, si2024freeu}. This motivates us to perform granular concept composition which is both frequency subband and timestep adaptive.

%% file: Sec/3_method.tex
\vspace{-0.3cm}
\section{MultLFG: Training-free Multi-LoRA Composition via Frequency-Guidance and Adaptive weighting}
\vspace{-0.3cm}

In this section, we describe our proposed algorithm, MultLFG for training-free frequency-aware multi-LoRA merging. The key idea is to decompose LoRA-based noise predictions into frequency subbands and perform adaptive merging based on relevance scores. Our approach leverages wavelet-domain representations, and adaptive weighting to improve compositionality and minimize concept interference.

\vspace{-0.3cm}
\paragraph{Problem Setup.} Let $\mathbf{z}_T$ be the initial latent, $\mathbf{c}$ the prompt embedding, $\mathcal{S}$ the noise scheduler, and $\{\mathcal{L}_i\}_{i=1}^N$ the set of LoRA adapters, each encoding a different concept. Our goal is to adaptively merge their outputs across denoising timesteps $t \in \{T, T{-}1, \dots, 1\}$ to generate a coherent output image.

\vspace{-0.3cm}
\paragraph{Step 1: Per-LoRA Noise Prediction.} For each timestep $t$, we activate each LoRA $\mathcal{L}_i$ and compute conditional and unconditional noise estimates. We also cache noise residuals from the previous step.
\begin{align*}
\epsilon_{\text{uncond}}^{(i,t)}, \epsilon_{\text{cond}}^{(i,t)} \gets \text{UNet}(\mathbf{z}_t, t, \mathbf{c}; \mathcal{L}_i), \quad 
\epsilon_{\text{uncond}}^{(i,t-1)}, \epsilon_{\text{cond}}^{(i,t-1)} \gets \text{UNet}(\mathbf{z}_t, t-1, \mathbf{c}; \mathcal{L}_i).
\end{align*}
Now from the noise, we obtain the clean latent, ($\alpha_{t}$, $\beta_{t}$ are scaling factors)
\begin{align*}
    z_{\text{uncond}}^{i,t} = (z_0 - \beta_{t}\epsilon_{\text{uncond}}^{(i,t)})/\alpha_{t}, \quad
    z_{\text{uncond}}^{i,t-1} = (z_0 - \beta_{t}\epsilon_{\text{uncond}}^{(i,t-1)})/\alpha_{t}
\end{align*}
Moreover, we also obtain the clean image for the next steps,
\begin{align*}
    x^{i,t} = \text{VAE}_{\text{decode}}(z_{\text{uncond}}^{i,t}), \quad
    x^{i,t-1} = \text{VAE}_{\text{decode}}(z_{\text{uncond}}^{i,t-1})
\end{align*}


\begin{figure}
    \centering
    \includegraphics[scale=0.35]
    {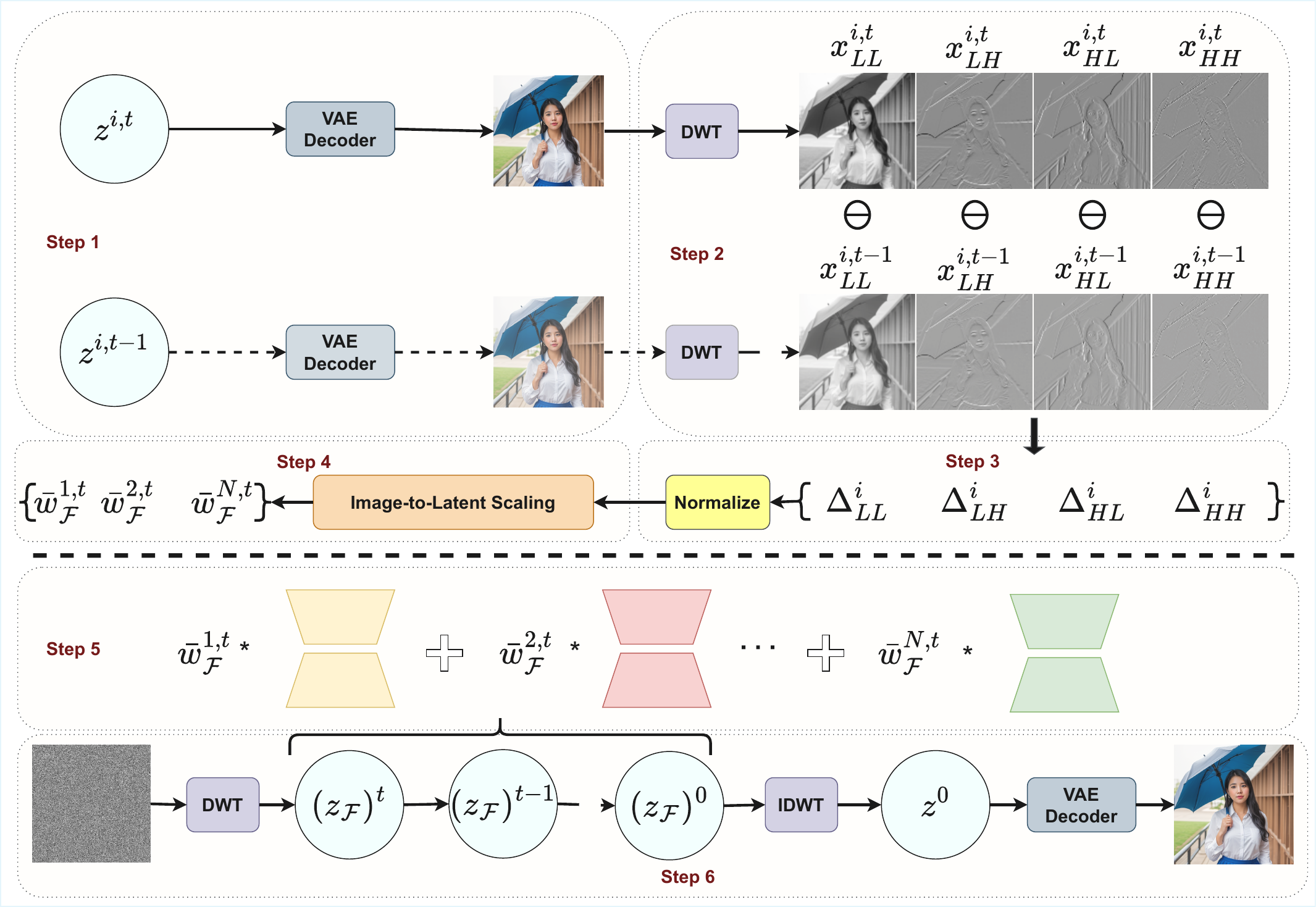}
    \caption{\small{Overview of MultLFG. (1) Per-LoRA noise is predicted from current and previous timesteps. (2) DWT is performed on denoised image and latent, (3) Temporal differences across consecutive timesteps are calculated followed by normalization by concept area, (4) Changes in images are scaled to changes in latent, (5) Adpative weights are computed based on importance of top-k LoRA, (6) These weights guide the weighted multi-LoRA composition in the wavelet domain for frequency-based guidance. The final image is generated by applying the IDWT and VAE decoding.}}
    \label{fig:main_arch}
    \vspace{-0.4cm}
\end{figure}

\vspace{-0.3cm}
\paragraph{Step 2: Frequency Decomposition.} 
Instead of applying guidance in the spatial domain—which often results in concept mixing and leakage (Fig.~\ref{fig:teaser}), we decompose the latent (or image) representation into multiple frequency subbands.
Specifically, we use DWT~\cite{mallat1989theory} for it's multi-scale subband decomposition, which enable finer control over concept composition.
We apply a 2D DWT to decompose both clean latent and decoded images into subbands, 
\begin{align*}
\{z_{\text{uncond}, LL}^{(i)}, z_{\text{uncond}, HL}^{(i)}, z_{\text{uncond}, LH}^{(i)}, z_{\text{uncond}, HH}^{(i)}\} = \text{DWT}(z_{\text{uncond}}^{(i)}).\hspace{2cm}\\
\{z_{\text{cond}, LL}^{(i)}, z_{\text{cond}, HL}^{(i)}, z_{\text{cond}, LH}^{(i)}, z_{\text{cond}, HH}^{(i)}\} = \text{DWT}(z_{\text{cond}}^{(i)}).\hspace{3cm}\\
\{x_{LL}^{(i,t)}, x_{HL}^{(i,t)}, x_{LH}^{(i,t)}, x_{HH}^{(i,t)}\} = \text{DWT}(x^{i,t}), \hspace{0.2cm}
\{x_{LL}^{(i,t-1)}, x_{HL}^{(i,t-1)}, x_{LH}^{(i,t-1)}, x_{HH}^{(i,t-1)}\} = \text{DWT}(x^{(i,t-1)})
\end{align*}
Each subband represents a localized frequency component. Next, we compute the adaptive weights for each subband to measure the concept’s activation strength in that frequency subband.



\vspace{-0.3cm}
\paragraph{Step 3: Temporal difference and normalization.}

Our intuition is that the contribution of each LoRA in a given timestep is proportional to the change it undergoes relative to the previous timestep—that is, we exploit the temporal cache produced by each LoRA.
Therefore, we calculate the difference in each subband in the image domain to measure the changes in subsequent timesteps in images. (refer to Fig.~\ref{fig:main_arch}, step 3) This provides the temporal change in image in a particular subband at a specific timestep.
\begin{align*}
    \Delta_{\text{img},\mathcal{F}}^{(i)} = | x_{\mathcal{F}}^{(i,t)} - x_{\mathcal{F}}^{(i,t-1)}|, \quad \forall \mathcal{F} \in \{\mathrm{LL}, \mathrm{HL}, \mathrm{LH}, \mathrm{HH}\}
\end{align*}
\vspace{-0.2cm}
However, the change in spatial dimensions may differ across concepts. For instance, concepts occupying larger areas (such as backgrounds) may result in greater absolute changes compared to spatially smaller concepts (e.g., small objects like bubble gum). Therefore, for each subband $\mathcal{F} \in \{LL, HL, LH, HH\}$, we normalize the changes in subsequent timesteps by dividing the area ($A$) of each concepts individually. The area of a concept is obtained by binary thresholding, followed by averaging w.r.t an area threshold. We set area threshold as, $\tau = 0.01$.
\begin{align*}
w_{\mathcal{F}}^{(i)} = \Delta_{img,\mathcal{F}}^{(i)} / A_{img,\mathcal{F}}^{(i)}, \quad
A_{img,\mathcal{F}}^{(i)} = \sum {\Delta_{img,\mathcal{F}}^{(i)}}^2 . \mathbf{1}({\Delta_{img,\mathcal{F}}^{(i)}}^2 \geq \tau)
\end{align*}
\vspace{-0.3cm}
\paragraph{Step 4: Image-to-latent scaling.}
At each timestep, we obtain how much each of the concepts vary in image-space in each wavelet subbands. However, we need to combine the scores in the latent space to guide the denoising process. Thus, we map the changes in image space to changes in latent. We perturb each sub-band of the image and measure the corresponding change in the latent space. Next, with this obtained relationship, we combine $w_{\mathcal{F}}^{(i)}$ to obtain the relative weightage of different concepts in the latent space $\bar{w}_{\mathcal{F}}^{(i)}$. We transform image-space perturbations into latent-space perturbations via the following steps:

\begin{enumerate}[leftmargin=*]
    \item Initially, we form the complete image space wavelet-domain perturbation vector,
    $(\Delta x_{LL}, \Delta x_{HL}, \Delta x_{LH}, \Delta x_{HH})$,
    which represents the changes in each of the four subbands.
    \item Then we reconstruct the corresponding image-space perturbation by applying IDWT: $\Delta (x_{\text{image}})_\mathcal{F} = \text{IDWT}(\Delta {x}_\mathcal{F}).$
    \item Next, we propagate this image-space perturbation for a particular sub-band $\mathcal{F}$ through the pretrained encoder ($E$) to obtain the latent-space change via a finite-difference approximation. We set, $\epsilon=10^{-5}$. This can be considered as a proxy to find the gradients w.r.t $E$.
    \[
    \Delta z_\mathcal{F} \approx \frac{E(x + \Delta (x_{\text{image}})_{\mathcal{F}}) - E(x)}{\epsilon} 
    \]
    $\Delta z_\mathcal{F}$, $\Delta (x_{\text{image}})_{\mathcal{F}}$ denotes the corresponding change in the whole latent space $z$, and image $x_{\text{image}}$ due to a small perturbation in the $\mathcal{F}$ sub-band $\Delta (x_{\text{image}})_{\mathcal{F}}$ in the image.  
    \item Then we decompose the overall change in latent representation $\Delta z_\mathcal{F}$ into DWT components, yielding:
    \[
    (\Delta z_\mathcal{F})_\mathcal{F'} = \text{DWT}(\Delta z_\mathcal{F}), \quad \text{for } \mathcal{F},\mathcal{F'} \in \{\mathrm{LL}, \mathrm{HL}, \mathrm{LH}, \mathrm{HH}\}.
    \]
    \item For a particular sub-band $\mathcal{F'}$, the relative importance of different concepts $\bar{w}_{\mathcal{F'}}$ can be expressed as,
\[
    \bar{w}_{\mathcal{F'}} = \sum_{\mathcal{F} \in \{\mathrm{LL}, \mathrm{HL}, \mathrm{LH}, \mathrm{HH}\}} w_{\mathcal{F}} * (\Delta z_\mathcal{F})_\mathcal{F'}
    \]
\end{enumerate}
This procedure transforms the image-space perturbations in each wavelet subband into the corresponding latent-space perturbations, thereby enabling effective latent guidance based on the individual subband contributions. 

\begin{algorithm}[t]
\caption{Adaptive Multi-LoRA Composition via Frequency-Guided Wavelet Analysis}
\label{alg:multi-lora-wavelet}
\KwIn{Initial latent $\mathbf{z}_T$, prompt embedding $\mathbf{c}$, LoRA adapters $\{\mathcal{L}_i\}_{i=1}^{N}$, noise scheduler $\mathcal{S}$, timesteps $T$, guidance scale $s$, top-$k$ selection}
\KwOut{Generated image $\mathbf{x}$}

Initialize latent: $\mathbf{z}_t \gets \mathbf{z}_T$ at timestep $t=T$\;

\For{$t = T, T-1, \dots, 1$}{

    \For{each adapter $\mathcal{L}_i$}{
        Set active adapter $\mathcal{L}_i$\;

        \textbf{// Step 1:} Compute noise predictions:\
        $\epsilon_{\text{uncond}}^{(i,t)}, \epsilon_{\text{cond}}^{(i,t)} \gets \text{UNet}(\mathbf{z}_t, t, \mathbf{c}; \mathcal{L}_i)$\;

        Compute previous-step noise:\
        $\epsilon_{\text{uncond}}^{(i,t-1)}, \epsilon_{\text{cond}}^{(i,t-1)} \gets \text{UNet}(\mathbf{z}_t, t-1, \mathbf{c}; \mathcal{L}_i)$\;

        Estimate clean latents:\
        $z_{\text{uncond}}^{(i,t)} \gets (z_0 - \beta_t\epsilon_{\text{uncond}}^{(i,t)})/\alpha_t$\; $z_{\text{uncond}}^{(i,t-1)} \gets (z_0 - \beta_t\epsilon_{\text{uncond}}^{(i,t-1)})/\alpha_t$\;

        Decode clean images:\
        $x^{(i,t)} \gets \text{VAE}_{\text{decode}}(z_{\text{uncond}}^{(i,t)})$\; 
        $x^{(i,t-1)} \gets \text{VAE}_{\text{decode}}(z_{\text{uncond}}^{(i,t-1)})$\;

        \textbf{// Step 2:} Perform wavelet decomposition (DWT) on images:\
        $\{x_{\mathcal{F}}^{(i,t)}\}, \{x_{\mathcal{F}}^{(i,t-1)}\} \gets \text{DWT}(x^{(i,t)}), \text{DWT}(x^{(i,t-1)})$\;

        \textbf{// Step 3:} Compute temporal differences per subband:\
        $\Delta_{\text{img},\mathcal{F}}^{(i)} \gets |x_{\mathcal{F}}^{(i,t)} - x_{\mathcal{F}}^{(i,t-1)}|$\;

        Normalize by area:\
        $w_{\mathcal{F}}^{(i)} \gets \Delta_{\text{img},\mathcal{F}}^{(i)} \big/ \sum (\Delta_{\text{img},\mathcal{F}}^{(i)})^2 \mathbf{1}\{(\Delta_{\text{img},\mathcal{F}}^{(i)})^2 \geq \tau\}$\;

        \textbf{// Step 4:}  Compute latent-space perturbations:\
        $\bar{w}_{\mathcal{F}}^{(i)} \gets \text{Img2Latent}(w_{\mathcal{F}}^{(i)})$\;
        

    }

    \textbf{// Step 5:}

    \For{each subband $\mathcal{F}\in\{LL,HL,LH,HH\}$}{
        Select top-$k$ LoRAs based on $\bar{w}_{\mathcal{F}}^{(i)}$\;

        Compute adaptive weights via softmax:\
        $\bar{w}_{\mathcal{F}} \gets \text{softmax}(\{\bar{w}_{\mathcal{F}}^{(i)}\}_{i=1}^{k})$\;

        Aggregate latent subbands adaptively:\
        $z_{\mathcal{F}} \gets \sum_{i=1}^{k} \bar{w}_{\mathcal{F}}^{(i)}[z_{\text{uncond},\mathcal{F}}^{(i)} + s_{\mathcal{F}}(z_{\text{cond},\mathcal{F}}^{(i)} - z_{\text{uncond},\mathcal{F}}^{(i)})]$\;
    }

     \textbf{// Step 6:} Reconstruct latent via inverse DWT:\
    $\hat{z}_t \gets \text{IDWT}(z_{LL}, z_{HL}, z_{LH}, z_{HH})$\;

    Compute aggregated noise:\
    $\hat{\epsilon}_t \gets (z_0 - \alpha_t\hat{z}_t)/\beta_t$\;

    Update latent state:\
    $\mathbf{z}_{t-1} \gets \mathcal{S}(\mathbf{z}_t, \hat{\epsilon}_t, t)$\;
}
Decode final latent to image:\
$\mathbf{x} \gets \text{VAE}_{\text{decode}}(\mathbf{z}_0)$\;
\end{algorithm}
\vspace{-0.2cm}

\vspace{-0.2cm}
\paragraph{Step 5: Adaptive Merging.} 
In LoRA-composite~\cite{zhong2024multi}, the scores are averaged at each timestep, which is suboptimal because different concepts may contribute unequally at different timesteps. To address this, we propose an adaptive combination scheme that enhances compositionality by dynamically weighting each concept. \textit{Our approach is motivated by the observation that the contribution of each LoRA in a given timestep is proportional to the change it undergoes relative to the previous timestep—that is, we exploit the temporal cache produced by each LoRA.}

For each subband $\mathcal{F} \in \{LL, HL, LH, HH\}$, we select the top-$k$ contributing LoRAs based on $\bar{w}^{(i)}_{\mathcal{F}}$ and compute softmax-normalized weights:
\begin{align*}
\bar{w}_{\mathcal{F}} &\gets \text{softmax}(\text{top-k}\{\bar{w}^{(i)}_{\mathcal{F}}|_{i=1,.., N}\}).
\end{align*}
The subband noise is then aggregated using weighted classifier-free guidance in wavelet domain:
\begin{align*}
z_{\mathcal{F}} = \sum_{i=1}^{k} \bar{w}_{\mathcal{F}}^{(i)} \left[ z_{\text{uncond},\mathcal{F}}^{(i)} + s_{\mathcal{F}} (z_{\text{cond},\mathcal{F}}^{(i)} - z_{\text{uncond},\mathcal{F}}^{(i)}) \right].
\end{align*}

\vspace{-0.2cm}
\paragraph{Step 6: Reconstruction and Update.} The reconstruced clean latent for timestep $t$ is obtained via inverse DWT, $\hat{z_t} = \text{IDWT}(z_{LL}, z_{HL}, z_{LH}, z_{HH}).$
Then we add noise to generate the noisy latent, $\hat{\epsilon_t} = (z_0 - \alpha_{t}\hat{z_t})/\beta_{t}.$
The latent is updated using the scheduler, $\mathbf{z}_{t-1} \gets \mathcal{S}(\mathbf{z}_t, \hat{\epsilon_t}, t).$
We cache the current predictions for use in the next timestep.

After all denoising steps, the output image is decoded from the final latent using the pretrained VAE, $\mathbf{x} \gets \text{VAE}_{\text{decode}}(\mathbf{z}_0)$.
This integrated framework of frequency-aware merging and adaptive weighting improves multi-concept compositionality and semantic faithfulness.
The overall method has been explained in Algorithm.~\ref{alg:multi-lora-wavelet}.

%% file: Sec/theory_5.tex
\vspace{-0.3cm}
\section{Theoretical and intuitive justification}
\vspace{-0.3cm}

\begin{lemma}[Frequency decomposition reduces interference]
Let \( x \) represent an image composed of multiple concepts \( \{c_i\}_{i=1}^n \). Consider wavelet decomposition of the image into frequency-specific subbands \( b \in \{LL, LH, HL, HH\} \):
$x = \sum_{b \in \{LL,LH,HL,HH\}} x_b.$

Define interference between concepts \( c_i \) and \( c_j \) within a subband \( b \) as:
\[
I_b(c_i, c_j) = \frac{\langle x_b^{c_i}, x_b^{c_j} \rangle}{\|x_b^{c_i}\|_2 \|x_b^{c_j}\|_2}.
\]
Then, interference in the frequency domain is strictly lower than the spatial domain interference:
\[
|I_b(c_i, c_j)| \leq |I_{\text{spatial}}(c_i, c_j)|,
\]
with equality only if \( x_b^{c_i} \) and \( x_b^{c_j} \) perfectly align spatially and frequency-wise.
\end{lemma}

\textit{Proof.} Proof is provided in the Supplementary. 
Intuitively, concept interference in multi-concept image generation occurs when attributes from different concepts overlap or conflict, leading to visual artifacts like misplaced elements or unintended blending (e.g., tie vanish/color change in Fig.~\ref{fig:teaser}). This problem is more prominent in spatial domain approaches, where attribute overlap can cause entangled representations (e.g., LoRA-composite, switch~\cite{zhong2024multi}).

Frequency domain guidance mitigates this issue by decomposing images/concepts into distinct frequency components (Fig.~\ref{fig:main_arch}), enabling more targeted control over individual attributes. Adaptive weights are applied to different frequency bands, allowing models to emphasize or suppress specific attributes and reduce interference. Low-frequency bands capture general shapes and backgrounds, while high-frequency bands represent fine details like textures and edges~\cite{mallat1989theory}. By manipulating these components separately, it's possible to maintain the integrity of each concept better (compared to spatial mixing) within the composite image. Prior work, such as WDIG~\cite{zhu2023wdig} and frequency-aware diffusion models~\cite{xiao2024frequency}, has shown that frequency domain processing can preserve image details and reduce artifacts associated with concept blending, resulting in improved image quality and content consistency.

%% file: Sec/4_experiments.tex
\vspace{-0.3cm}
\section{Experiments}
\vspace{-0.3cm}
\begin{figure*}
    \centering
    \includegraphics[width=0.85\linewidth]{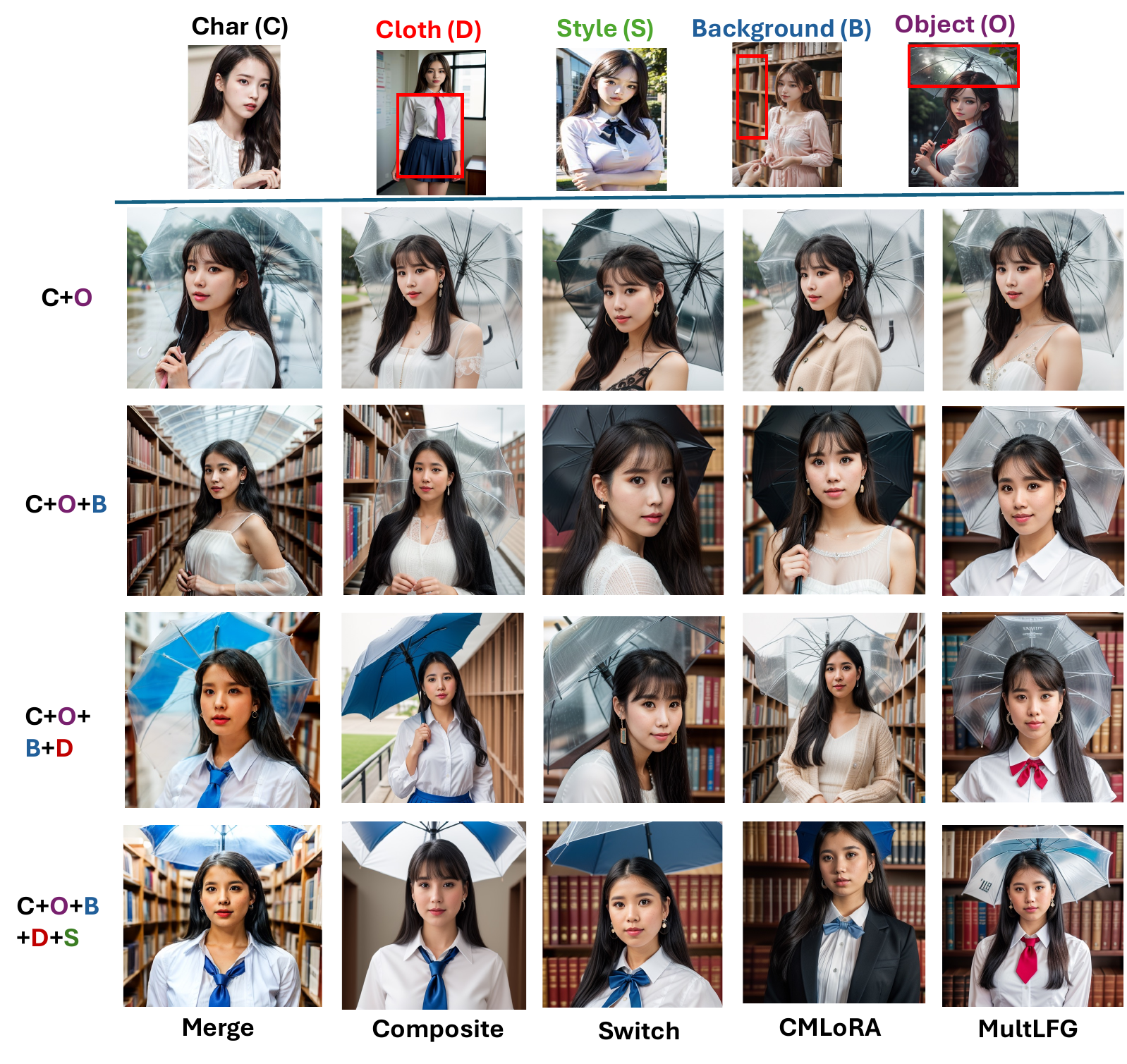}
    \vspace{-0.4cm}
    \caption{\small{Comparison on multi-LoRA composition for realistic images. We observe concept mix (color of tie exchanged with color of skirt in 4th, 5th row), or concept vanish (tie/ dress is missing in 4th row) in baselines, whereas MultLFG (last column) minimizes concept mix or vanish, while maintaining quality.}}
    \label{fig:compare_reality}
    \vspace{-0.7cm}
\end{figure*}

\textbf{Experimental setup.}
Experiments are conducted using the ComposeLoRA testbed~\cite{zhong2024multi}, consisting of 22 LoRAs across realistic and anime styles, covering characters, clothing, styles, backgrounds, and objects. The evaluation involves 480 unique sets, categorized by the number of LoRAs (2, 3, 4, 5) with each set including one character LoRA to ensure no category overlap. Additional details are provided in the supplementary. To the best of our knowledge, we are not aware of any other publicly available benchmark to evaluate this particular task.

\textbf{Baselines.}
We compare with the following baselines: (1) Naive~\cite{rombach2022high}: images generated through only prompts without using concept LoRAs. (2) Merge~\cite{zhong2024multi}: merging the concept LoRA weights by simple addition, (3) Switch~\cite{zhong2024multi} : Switch across concept LoRAs across timesteps, (4) Composite~\cite{zhong2024multi} : averaging concept LoRA scores across timesteps, (5) LoRAHub~\cite{huang2023lorahub} : Combine multiple LoRA with few-shot learning, (6) CMLoRA~\cite{zou2025cached}: combine LoRA based on fourier-based caching method.

\textbf{Evaluation metric.}
To evaluate concept composition, we use CLIP-score~\cite{zou2025cached} to assess text-image alignment and GPT-4V for detailed composition and quality evaluation as designed by~\cite{zou2025cached}. GPT-4V scores generated images based on composition and overall quality, ranging from 0 to 10, guided by specific prompts. GPT-4V evaluation is done pairwise (baselines vs MultLFG), and avg. score w.r.t the baselines are reported in Tab.~\ref{tab:quality_gpt4_combined} and Tab.~\ref{tab:composition_gpt4_combined}. More details are in supplementary.


\textbf{Implementation details.}
In our experiments, we employ SD-v1.5~\cite{rombach2022high} and SDXL~\cite{podell2023sdxl} as the base model. For the realistic style subset, we set the model to 100 denoising steps with a guidance scale of 7 and a resolution of 512x512. For the anime style subset, we increase the denoising steps to 200 with a guidance scale of 10 while maintaining the same resolution. All experiments are conducted on a single A5000 GPU with 24GB of memory. The DWT is implemented using `Haar' wavelets~\cite{mallat1989theory}.

\begin{figure*}
    \centering
    \includegraphics[width=0.9\linewidth]{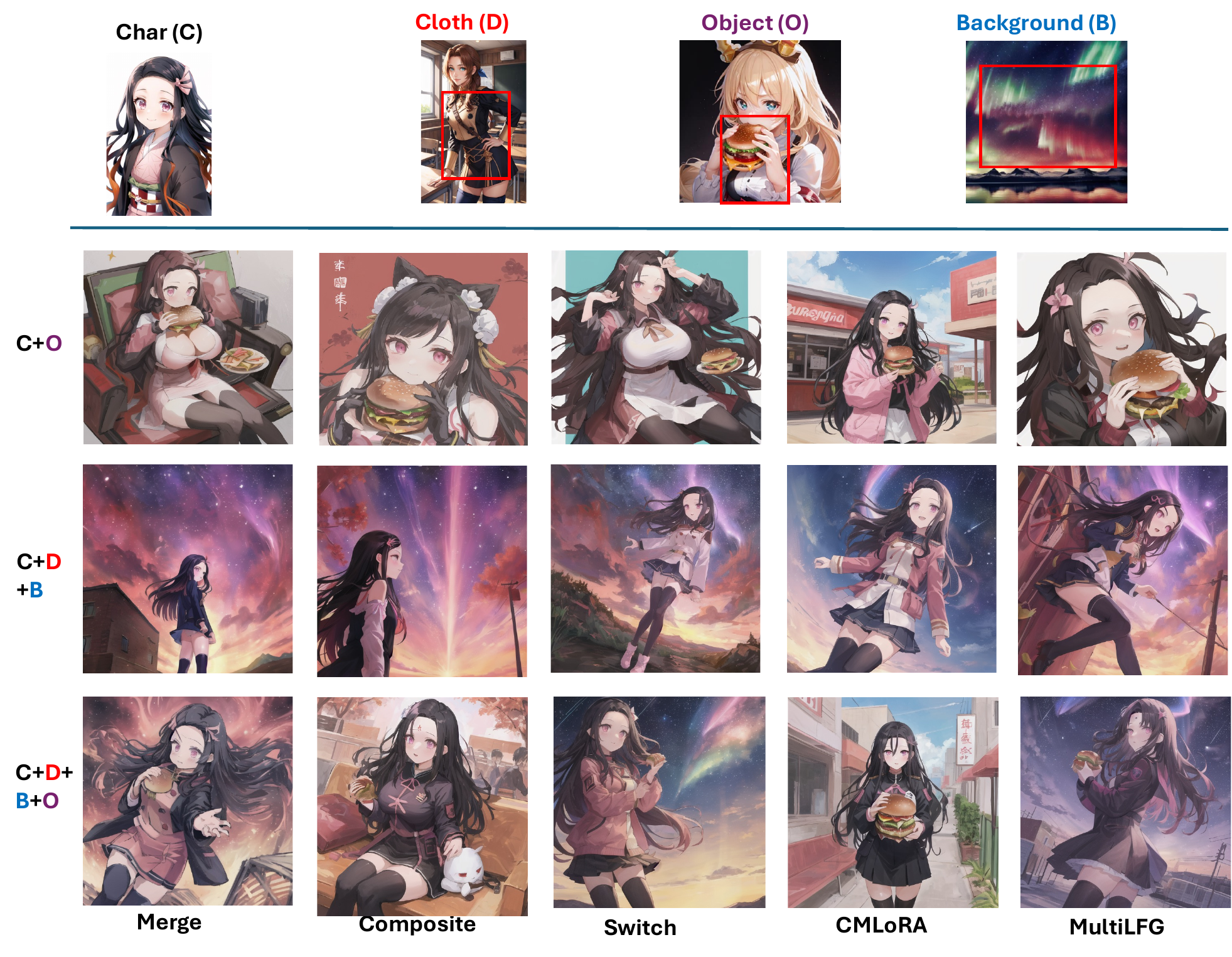}
    \vspace{-0.4cm}
    \caption{\small{Comparison for anime images. Implausible concept composition (burger is floating in 2nd row baselines), or concept erasure (different dress in 3rd row, either dress or background inconsistent in 4th row) happens in baselines. MultLFG combines concepts better while maintaining quality (last column).}}
    \label{fig:compare_cartoon}
    \vspace{-0.5cm}
\end{figure*}

\textbf{Quantitative and qualitative results.}
We present CLIP-score comparisons with baselines for realistic and anime styles in Tab.~\ref{tab:clip_score_combined}, where MultLFG consistently outperforms the baselines. Additionally, compositionality and image quality evaluations using GPT-4 are reported in Tab.~\ref{tab:composition_gpt4_combined} and Tab.~\ref{tab:quality_gpt4_combined}, further demonstrating MultLFG’s superior performance across both dimensions. Qualitative comparisons are illustrated in Fig.~\ref{fig:compare_reality} and Fig.~\ref{fig:compare_cartoon}.

\textbf{Ablations.}
We conduct ablation studies on the proposed components for realistic images in Tab.~\ref{tab:ablation_reality}, demonstrating that both frequency-guidance and adaptive merging effectively enhance concept composition. Additionally, we evaluate the impact of k in top-k merging, as presented in Tab.~\ref{tab:topk_ablation}.


\textbf{User study.} To account for potential unreliability in perceptual metrics, we conduct a human preference study using Amazon Mechanical Turk (AMT) to evaluate compositionality and image quality. Fifty unbiased users ranked our method against baseline approaches in 1,000 questionnaires. The aggregated responses in Table~\ref{tab:user_study} indicate that MultLFG substantially outperformed the baselines. Further details are provided in the supplementary material.


\textbf{Complexity analysis.}
We perform complexity analysis in terms of inference time in Tab.~\ref{tab:complexity}. We observe that MultLFG improves the concept composition with increased inference time. 

\textbf{Limitations.}
While MultLFG enhances both compositionality and quality; it is comparatively slower than the baselines, and optimizing its efficiency will be the focus of future work.


\vspace{-0.5cm}
\begin{table}[h!]
\centering
\begin{minipage}{0.25\textwidth}
\centering
\caption{\small{User study}}
\scalebox{0.7}{
\begin{tabular}{l|c}
\hline
\textbf{Method} & \textbf{User preference (\%)} \\ \hline
Naive & 2.0 \\ 
Merge & 7.0 \\ 
Switch & 18.0\\ 
Composite & 13.0 \\ 
LoRAHub & 16.0\\ 
CMLoRA & 19.0\\ 
\ourcolor MultLFG & \ourcolor 25.0\\ \hline 
\end{tabular}}
\label{tab:user_study}
\end{minipage}
\hfill
\begin{minipage}{0.7\textwidth}
\centering
\caption{\small{CLIP-Score across different compositional levels for Realistic and Anime. (N = No. of concepts)}}
\scalebox{0.65}{
\begin{tabular}{l|c|c|c|c|c|c|c|c}
\hline
\textbf{Method} & \multicolumn{4}{c|}{\textbf{Realistic}} & \multicolumn{4}{c}{\textbf{Anime}} \\ \hline
 & \textit{N=2} & \textit{N=3} & \textit{N=4} & \textit{N=5} & \textit{N=2} & \textit{N=3} & \textit{N=4} & \textit{N=5} \\ \hline
Naive~\cite{rombach2022high} (CVPR'22) & 35.04 & 34.92 & 34.38 & 33.80 & 35.07 & 34.89 & 34.27 & 32.77 \\ 
Merge~\cite{zhong2024multi} (TMLR'24) & 33.72 & 34.13 & 33.39 & 32.36 & 35.13 & 35.42 & 34.16 & 32.63 \\ 
Switch~\cite{zhong2024multi} (TMLR'24) & 35.39 & 35.10 & 34.97 & 33.47 & 35.28 & 35.48 & 34.53 & 34.14 \\ 
Composite~\cite{zhong2024multi} (TMLR'24) & 35.07 & 34.08 & 34.80 & 32.58 & 34.34 & 34.37 & 34.16 & 32.93 \\ 
LoRAHub~\cite{huang2023lorahub} (COLM'24) & 35.68 & 35.12 & 34.77 & 33.48 & 35.31 & 35.52 & 34.47 & 33.88 \\ 
CMLoRA~\cite{zou2025cached} (ICLR'25) & 35.42 & 35.21 & 35.20 & 34.34 & 35.55 & 35.55 & 35.79 & 35.69 \\ 
\ourcolor MultLFG & \ourcolor \textbf{36.42} & \ourcolor \textbf{36.12} & \ourcolor \textbf{35.91} & \ourcolor \textbf{35.62} & \ourcolor \textbf{36.72} & \ourcolor \textbf{36.13} & \ourcolor \textbf{36.45} & \ourcolor \textbf{36.22} \\ \hline
\end{tabular}}
\label{tab:clip_score_combined}
\end{minipage}
\vspace{-0.6cm}
\end{table}

\begin{table}[h!]
\centering
\begin{minipage}{0.46\textwidth}
    \centering
    \caption{\small{Compositional Quality using GPT4v Evaluation on Realistic and Anime Images}}
    \scalebox{0.63}{
    \begin{tabular}{l|c|c|c|c|c|c|c|c}
    \hline
    \textbf{Method} & \multicolumn{4}{c|}{\textbf{Realistic}} & \multicolumn{4}{c}{\textbf{Anime}} \\ \hline
     & \textit{N=2} & \textit{N=3} & \textit{N=4} & \textit{N=5} & \textit{N=2} & \textit{N=3} & \textit{N=4} & \textit{N=5} \\ \hline
    Naive & 7.31 & 6.92 & 5.45 & 4.89 & 7.51 & 6.45 & 5.11 & 4.25 \\ 
    Merge & 8.21 & 7.11 & 6.02 & 5.25 & 8.02 & 7.06 & 5.65 & 4.89 \\ 
    Switch & 8.62 & 7.53 & 6.92 & 6.12 & 8.32 & 7.56 & 6.03 & 5.32 \\ 
    Composite & 8.47 & 7.47 & 6.88 & 6.05 & 8.20 & 7.38 & 5.88 & 5.15 \\ 
    LoRAHub & 8.39 & 7.44 & 6.81 & 6.07 & 8.18 & 7.35 & 5.82 & 5.10 \\ 
    CMLoRA & 8.65 & 7.58 & 6.98 & 6.15 & 8.31 & 7.50 & 6.02 & 5.23 \\ 
    \ourcolor MultLFG & \ourcolor \textbf{8.88} & \ourcolor \textbf{8.02} & \ourcolor \textbf{7.22} & \ourcolor \textbf{6.32} & \ourcolor \textbf{8.82} & \ourcolor \textbf{7.95} & \ourcolor \textbf{6.49} & \ourcolor \textbf{5.69} \\ \hline
    \end{tabular}}
    \label{tab:composition_gpt4_combined}
\end{minipage}%
\hfill
\begin{minipage}{0.46\textwidth}
    \centering
    \caption{\small{Image Quality using GPT4v Evaluation on Anime and Realistic Images}}
    \scalebox{0.63}{
    \begin{tabular}{l|c|c|c|c|c|c|c|c}
    \hline
    \textbf{Method} & \multicolumn{4}{c|}{\textbf{Anime}} & \multicolumn{4}{c}{\textbf{Realistic}} \\ \hline
     & \textit{N=2} & \textit{N=3} & \textit{N=4} &\textit{ N=5} & \textit{N=2} & \textit{N=3} & \textit{N=4} &\textit{ N=5} \\ \hline
    Naive & 7.23 & 7.49 & 7.80 & 7.77 & 7.52 & 8.00 & 7.82 & 6.88 \\ 
    Merge & 7.92 & 8.02 & 8.23 & 8.22 & 8.05 & 8.56 & 8.23 & 7.39 \\ 
    Switch & 8.89 & 8.66 & 8.92 & 8.77 & 8.64 & 9.10 & 8.92 & 8.48 \\ 
    Composite & 9.05 & 8.97 & 9.03 & 8.98 & 8.95 & 9.22 & 9.07 & 8.97 \\ 
    LoRAHub & 9.00 & 8.92 & 8.97 & 8.82 & 8.87 & 9.18 & 8.92 & 8.85 \\ 
    CMLoRA & 9.12 & 9.03 & 9.10 & 9.05 & 8.98 & 9.25 & 9.10 & 9.01 \\ 
    \ourcolor MultLFG & \ourcolor \textbf{9.71} & \ourcolor \textbf{9.55} & \ourcolor \textbf{9.54} & \ourcolor \textbf{9.48} & \ourcolor \textbf{9.43} & \ourcolor \textbf{9.52} & \ourcolor \textbf{9.45} & \ourcolor \textbf{9.37} \\ \hline
    \end{tabular}}
    \label{tab:quality_gpt4_combined}
\end{minipage}
\vspace{-0.8cm}
\end{table}

\begin{table}[h!]
\centering
\begin{minipage}{0.4\textwidth}
    \centering
    \caption{\small{Ablation of components}}
    \scalebox{0.7}{
    \begin{tabular}{l|c|c|c|c}
    \hline
    \textbf{Method} & \textbf{\textit{N=2}} & \textbf{\textit{N=3}} & \textbf{\textit{N=4}} & \textbf{\textit{N=5}} \\ \hline
    Composite & 35.07 & 34.08 & 34.80 & 32.58 \\ 
    + Freq. guidance & 35.98 & 35.46 & 35.25 & 34.87 \\ 
    + adaptive weight & 36.42 & 36.12 & 35.91 & 35.62 \\ \hline
    \end{tabular}}
    \label{tab:ablation_reality}
\end{minipage}%
\hfill
\begin{minipage}{0.3\textwidth}
    \centering
    \caption{\small{Inference-time(s)}}
    \scalebox{0.65}{
    \begin{tabular}{l|c|c|c|c}
    \hline
    \textbf{Method} & \textbf{\textit{N=2}} & \textbf{\textit{N=3}} & \textbf{\textit{N=4}} & \textbf{\textit{N=5}} \\ \hline
    Merge & 20 & 21 & 22 & 24 \\ 
    Switch & 16 & 18 & 19 & 20 \\ 
    Composite & 60 & 70 & 76 & 90 \\ 
    MultLFG & 90 & 140 & 180 & 230 \\ \hline
    \end{tabular}}
    \label{tab:complexity}
\end{minipage}%
\hfill
\begin{minipage}{0.25\textwidth}
    \centering
    \caption{\small{Top-k Ablation}}
    \scalebox{0.63}{
    \begin{tabular}{l|c|c|c|c}
    \hline
    \textbf{K} & \textbf{\textit{N=2}} & \textbf{\textit{N=3}} & \textbf{\textit{N=4}} & \textbf{\textit{N=5}} \\ \hline
    1 & 36.21 & 36.07 & 35.75 & 35.48 \\ 
    2 & 36.42 & 36.12 & 35.91 & 35.50 \\ 
    3 & - & 36.08 & 35.68 & 35.62 \\ 
    4 & - & - & 35.71 & 35.55 \\ 
    5 & - & - & - & 35.39 \\ \hline
    \end{tabular}}
    \label{tab:topk_ablation}
\end{minipage}
\vspace{-0.5cm}
\end{table}

%% file: Sec/5_conclusion.tex
\vspace{-0.4cm}
\section{Conclusion}
\vspace{-0.3cm}
In this work, we introduced MultLFG, a novel framework for training-free multi-LoRA composition that leverages frequency-domain guidance for effective and adaptive merging of multiple LoRAs. By decomposing latent representations into distinct frequency bands and applying targeted guidance at each denoising timestep, MultLFG effectively mitigates concept mixing and leakage while enhancing compositional fidelity. Experimental evaluations on the ComposLoRA benchmark demonstrate that MultLFG significantly outperforms existing baselines in both compositional quality and image fidelity across diverse styles and concept sets.  
